# A Simple Approach for Biometrics: Finger-Knuckle Prints Recognition Based on a Sobel Filter and Similarity Measures

É. O. Rodrigues, T. M. Porcino and Aura Conci*, Aristofanes C. Silva†, *Department of Computer Science, Universidade Federal Fluminense, Rua Edmundo March s/n - Boa Viagem, Niterói - RJ, 24435-030, Brazil.
†Department of Electrical Engineering, Universidade Federal do Maranhão, Av. dos Portugueses s/n - São Luiz - MA, 65085-580, Brazil.
*erickr@id.uff.br, tmp1986@gmail.com, aconci@ic.uff.br, ari@dee.ufma.br*

***Abstract*—The objective of this work is to propose a novel methodology for the finger knuckle print recognition, which is essentially a digital photo of the finger-knuckle region. We have employed very simple concepts of visual computing such as a filter based on the Sobel operator for finding edges and a simple noise reduction algorithm. These operations are exceptionally fast and produce binary images, which are very efficient to process and to store. Furthermore, alongside this preprocessing, some similarity measures were also regarded and evaluated for the task. After preprocessing an input finger it is compared to all the images of fingers in the dataset, one by one. We have obtained up to 17.02% of successful recognitions (true positive rate) with a large dataset.**

## I. Introduction

Biometric measurements are vastly used for recognizing individuals. These measures are commonly employed in security systems that usually allow or deny an action to a user depending on the outcome of the recognition [1], [2]. As usual in computer systems, the collected and analysed data are discrete samples, which in terms suppresses the quality of the recognition. Therefore, flaws are certainly present in these systems. However, reducing it at most would ensure more reliable recognitions.

The performance of the recognition between different systems and methodologies are often compared using indexes such as accuracy and the ones derived from confusion matrixes, where it is desired to minimize the rate of false positives and to maximize the accuracy of the system. The usability of the system and processing time are also important points to be regarded.

In this work, we propose a simple yet robust method for recognizing individuals by just using principles of computer vision. We work on finger-knuckle images, which is essentially a photo of the upper part of the finger. The recognition is then based on the biometrical information extracted from this area.

## II. Literature Review

Probably, the first publicly available database of finger-knuckles was provided by Zhang [3], where it was called Finger-Knuckle Prints (FKP). The dataset currently contains images from 147 volunteers at total. The acquisition was performed in 12 sections for the left and right index and middle fingers. The average time interval between the first and second sessions was about 25 days.

Le-qing [4] uses Speeded-Up Robust Features (SURF) to obtain keypoints in the images, combined with a Random Sample Consensus (RANSAC) strategy for matching the fingers, where they obtain an accuracy of 96.91% on the recognition of the middle finger.

Zhang et al. [5] proposed using a combination of local features, achieving a true positive rate of 15.15%. Furthermore, Guru et al. [6] employed moments and feature fusion to achieve an accuracy up to 92.24%.

All the previously addressed works proposed successful methods for the biometrical recognition of the regarded region. However, most of them are significantly complicated and may consume a considerable amount of processing time.

In this work, we propose a novel distinct approach, based mainly on a extension of the Sobel operator [7] and similarity measures. The Sobel operator is very simple, fast and is commonly used to detect edges in images. Furthermore, we also compare the three similarity measures we have evaluated and discuss their performance.

## III. Materials and Methods

The dataset used in our work was the one provided by Zhang et al. [3], as previously addressed. The authors provide both the entire image of the finger and their already extracted Region of Interest (ROI), which corresponds to the central knuckle-part of the finger. In this work, we have used the ROIs provided by them. Fig. 1 summarizes the steps of our methodology.

### A. Edge Detection

As it can be seen in Fig. 1, at first we apply two adaptations of the Sobel filter, the first one will detect valleys that were filled with shadows, while the second one detects valleys that were enlightened. Fig. 2 shows a possible input image and the result of the edge detection.

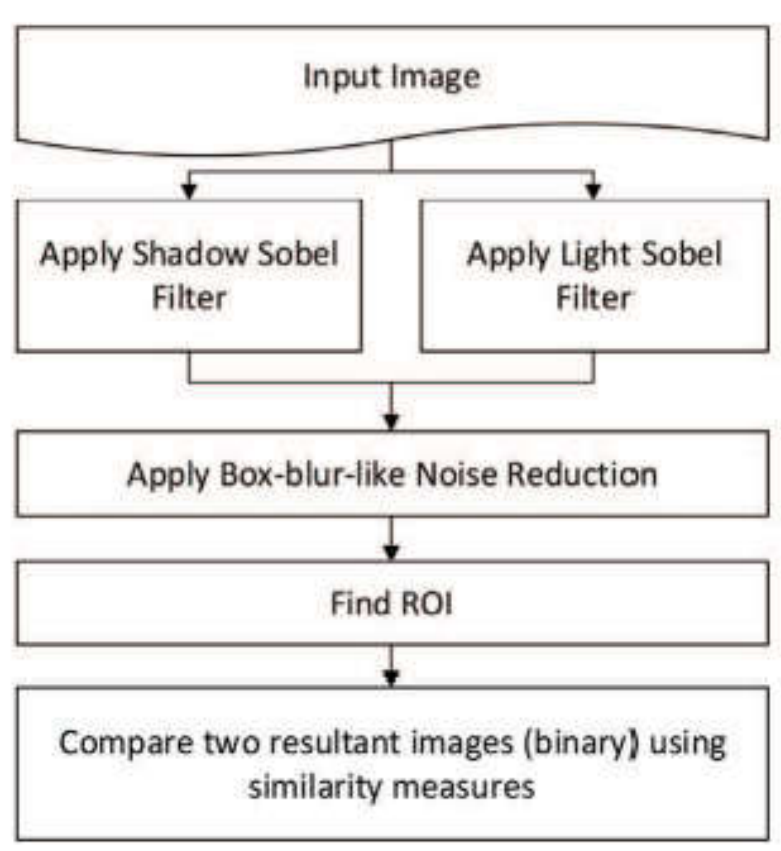

Figure 1: Steps of our proposed methodology.

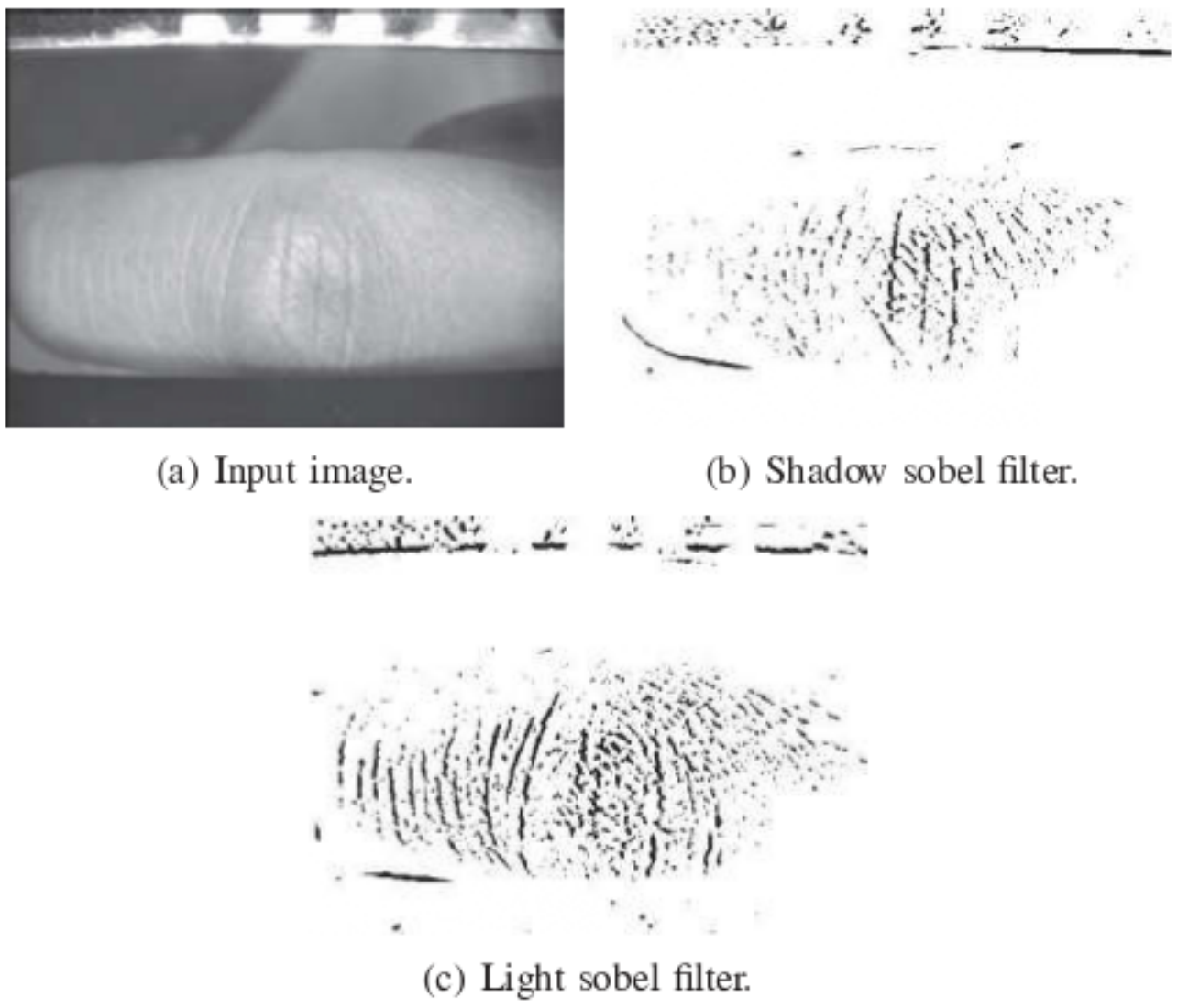

(a) Input image. (b) Shadow sobel filter.

(c) Light sobel filter.

Figure 2: Sobel-like applied filters.

The Shadow and Light Sobel filters are defined by Equations 1 and 3, respectively, where $P$ represents the image and $P_{i,j}$ a value of a pixel at line $i$ and column $j$, $t$ represents a threshold for the edge to be marked, and $d$ a distance parameter (in classical Sobel operators $d$ would be close to 1), and $\wedge$ represents the boolean *and* operation.

$$SS(P,d,t,i,j) = \begin{cases} 1, & if\ (s_1 \wedge s_2) \\ 0, & otherwise \end{cases} \quad (1)$$

where,

$$\begin{aligned} s_1 &= P_{i-d,j} - P_{i,j} > t \wedge P_{i+d,j} - P_{i,j} > t, \\ s_2 &= P_{i,j-d} - P_{i,j} > t \wedge P_{i,j+d} - P_{i,j} > t \end{aligned} \quad (2)$$

and

$$LS(P,d,t,i,j) = \begin{cases} 1, & if\ (l_1 \wedge l_2) \\ 0, & otherwise \end{cases} \quad (3)$$

where,

$$\begin{aligned} l_1 &= P_{i,j} - P_{i-d,j} > t \wedge P_{i,j} - P_{i+d,j} > t, \\ l_2 &= P_{i,j} - P_{i,j-d} > t \wedge P_{i,j} - P_{i,j+d} > t \end{aligned} \quad (4)$$

The main difference between the variables $s_{1..2}$ and $l_{1..2}$ is the orientation of the computation, which captures either light or shadow valleys in each occasion. We have tried to fuse these two generated images but the noise on the resultant image increased significantly, even if applying a noise reduction before fusing.

Furthermore, some important information is also lost on the fusion. A registration operation was also evaluated to fuse the images but it increases the processing time. Besides, the results were not visually pleasant and this could increase the error rate of the system, since the results of the registrations would be different for each specific case.

### *B. Noise Reduction*

The parameters used in this work (also to generate the images in Fig. 2) were $t = 8$ and $d = 4$ for the shadow image and $t = 5$ and $d = 4$ for the light image, where the resolution of the input images were $384 \times 288$. However, it is clear that both the shadow and light Sobel images contain a significant amount of noise. If we decrease the parameter $t$ then some important information such as the actual lines of the finger start to be ignored (not extracted). Therefore, we have employed a box-blur like noise reduction algorithm [8], to reduce this generated noise instead. The noise reduction is defined by Equation 5.

$$NR(P,t,i,j) = \begin{cases} 1, & if\ (\sum_{y=-a_y+i}^{a_y+i} \sum_{x=-a_x+j}^{a_x+j} P_{y,x}) > t \\ 0, & otherwise \end{cases} \quad (5)$$

What the noise reduction in Equation 5 essentially does is taking a pixel at position $(j, i)$ and evaluating a squared neighbourhood around it, checking whether this neighbourhood has at least $t$ pixels within that area. If that is so, then the central pixel at position $(j, i)$ remains painted. Otherwise it is erased. Obviously, this operation is performed for every pixel of the image, not just for a single position. In other words, the noise reduction operation essentially erases isolated pixels in the image. The variable $a_x$ represents the size of the area on the $x$ axis and $a_y$ on the $y$ axis.

We also perform an adaptation on the noise reduction algorithm by decrementing $a_x$ and $a_y$, until they reach 1. For every decrementation, the Equation 5 is called again. The parameter $t$ was empirically chosen to be 7 for the light image and 13 to the shadow image while $a_x$ and $a_y$ were both equal to 5 on the light image and 10 on the shadow image (initially, before the decrementations). Fig. 3 shows the images (b) and (c) of Fig. 2 after applying the noise reduction algorithm.

### *C. ROI Extraction*

Just after the sobel and noise reduction algorithm, we extract the ROI of the images. The ROI is set as a region of $220 \times 110$ pixels and is obtained as shown in [3]. So, after applying the preprocessing algorithms we get the extracted ROIs of the images (b) and (c), shown in Fig. 4. However, if desired, this step can be performed before the preprocessing instead.

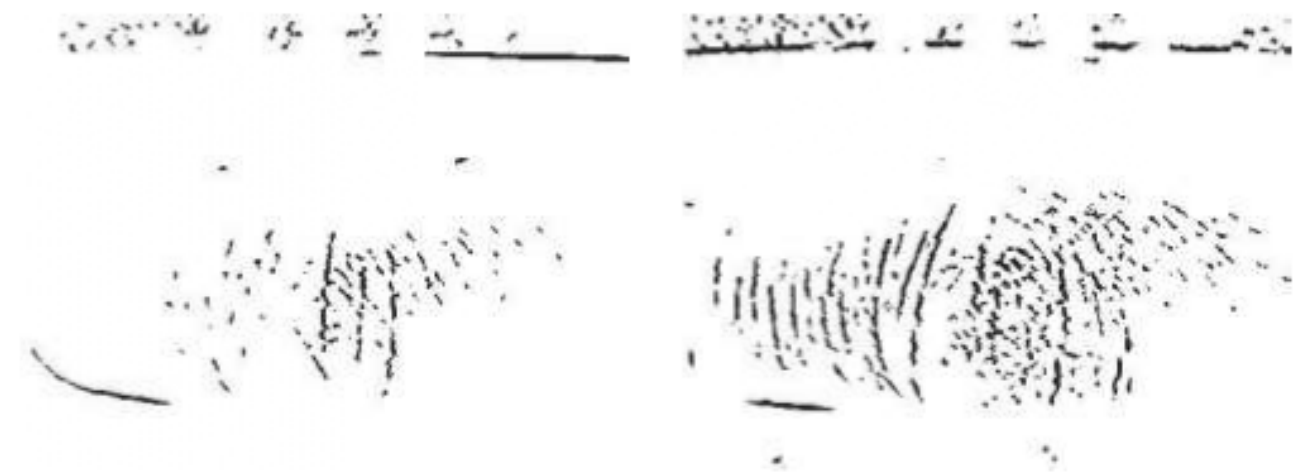

(a) Shadow image with noise reduction. (b) Light image with noise reduction.

Figure 3: Noise reduction algorithm.

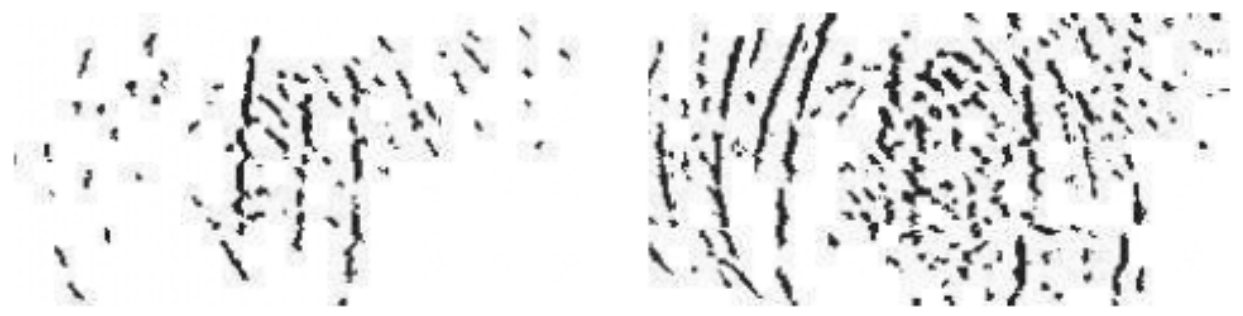

(a) ROI of the shadow image after noise reduction. (b) ROI of the light image after noise reduction.

Figure 4: ROI Extraction.

### D. Similarity Measures

Assuming we are to compare two fingers, both images of these two fingers are processed following the previously described steps. At first, we take the light and shadow binary images based on the Sobel operator. Later, we apply a noise reduction algorithm. From the images with less noise we then extract their ROIs. Thus, a single finger would have two ROIs, the shadow and light ROIs. These two ROIs are compared to the light and shadow ROIs of the second finger and the one that achieved the least distance in both images is said to be the expected subject to be recognized. In order to compare the images (i.e., the ROIs), we use some similarity measures.

The first similarity measure is the simplest one, which consists of summing the absolute differences between the pixel values [9]. In this case, since the images are binary, it can also be considered the Hamming distance [10]. The formulation for this Mean Absolute (MA) measure is given by Equation 6, where $P_1$ and $P_2$ represent the images being compared.

$$MA(P_1, P_2) = \frac{1}{hw} \sum_{y=0}^{h} \sum_{x=0}^{w} |P_{1_{y,x}} - P_{2_{y,x}}| \tag{6}$$

The downside of the MA measure is that it does not account much for similarity but for exact matches. Therefore, we have also tested the Hausdorff Distance (HD) [11] that, roughly speaking, accounts for the worst difference between the images being compared. The HD formulation is given by Equation 7, where $||.||$ represents the distance norm (e.g., euclidean distance), and $p_1$ and $p_2$ are points within the image or set $P_1$ and $P_2$, respectively.

$$HD(P_1, P_2) = max\{ \underset{p_1 \in P_1}{sup} \underset{p_2 \in P_2}{inf} ||p_1 - p_2||, \underset{p_2 \in P_2}{sup} \underset{p_1 \in P_1}{inf} ||p_1 - p_2||\} \tag{7}$$

Table I: Confusion matrixes for distinct similarities measures.

| Similarity Measure | TP (%) | TN (%) | FP (%) | FN (%) |
|---|---|---|---|---|
| Chamfer Distance | 17.0238 | 99.9877 | 82.9762 | 0.0123 |
| Hausdorff Distance | 2.6046 | 99.9855 | 97.3953 | 0.0144 |
| Mean Absolute | 0.7856 | 99.9852 | 99.2113 | 0.0148 |

The problem with HD is that it is very sensitive to outliers. If the images being compared are almost equal except for a noise point in one of them that is very far from the remaining image points, then HD would be huge (computed based on that outlier). In this work, we have applied a noise reduction algorithm to the images also to reduce this problem. It does not exempt HD from producing bad results in similar occasions but it certainly reduces it.

Still, we wanted a measure that accounts more for similarity rather than exact superposition or size of the error. The Chamfer Distance (CD) is the measure that does it better between these three. Given two images $P_1 = \{p_{k_1}\}_{k_1=1}^{n_1}$ and $P_2 = \{p_{k_2}\}_{k_2=1}^{n_2}$, CD is the mean of distances between each point $p_{k_1} \in P_1$ and its closest point in $P_2$ [12] as shown in Equation 8, where $\tau$ is a threshold value, reducing the effect of outliers and missing edges. In this work, we have considered $\tau = 2$.

$$CD(P_1, P_2) = \frac{1}{n_1} \sum_{p_{k_1} \in P_1} max(\underset{p_{k_2} \in P_2}{min} ||p_{k_1} - p_{k_2}||, \tau) \tag{8}$$

Suppose we need to identify an incoming finger, then the images (light and shadow) are compared to all the images (light and shadow) in the database. Among the fingers in the database, the one that obtained the minimum distance (minimum score) or maximum similarity (the images are theoretically more equal) is chosen as the right person.

## IV. Results

The final results are shown in Table I. The true positive (TP) index stands for the rate of correct recognitions and is the most important index in this problem. Due to the huge quantity of comparisons, the true negative index would be huge even if the TP rate is zero, and that influences positively on the accuracy. However, that does also influences negatively on the true positive index. In summary, using the Chamfer distance, we were able to successfully recognize the person assuming they are providing just a single sample or finger image on 17.02% of the comparisons.

For each candidate finger to be recognized, a total of 1763 comparisons are performed $(12 \times 147 - 1)$. Among these 1763 comparisons, just 12 comparisons stand for the right person (which is the number of images of each finger for each person on the database). Therefore, if the recognition was random, the rate of successfully recognizing one person would be next to 0.680%. And the rate of successfully recognizing all the users by chance in the database would be something highly uncommon, with a probability of approximately $1.1 \times 10^{-24}\%$. Therefore, achieving a successful recognition rate of 17.02% is very expressive in comparison.

Table II: Accuracy and time comparison.

| Similarity Measure | Accuracy (%) | Time (×) |
|---|---|---|
| Chamfer Distance | 99.9902 | 105.0234 |
| Mean Absolute | 99.9854 | 1 |
| Hausdorff Distance | 99.8887 | 223.557 |

Table III: Recognition rates per finger.

| Fingers | TP (%) | FN (%) | Accuracy (%) |
|---|---|---|---|
| Left Index | 30.7738 | 0.0412 | 99.9176 |
| Left Middle | 26.3690 | 0.0438 | 99.9123 |
| Right Index | 18.2142 | 0.0487 | 99.9026 |
| Right Middle | 26.4285 | 0.0438 | 99.9126 |

Table II shows the accuracies obtained with each similarity measure and the averaged processing time each similarity measure required related to the mean absolute measure, which was the fastest one. In other words, the values in the *Time* column stand for how many times slower were the similarity measures in relation to the mean absolute one.

In addition, Table III compares the results obtained using the Chamfer Distance when only left indexes, left middles, right indexes and right middles fingers are considered (not joined together as in the previously addressed rates).

Finally, Table IV compares our results to results obtained by related works on the same database. We have averaged the rates for all the fingers on the database.

## V. Conclusion

Our proposed approach is simple and robust. We have constrained ourselves to very simple techniques of computer vision. Besides, it can be highly optimized to run in low capable processing systems. For instance, we work with binary images and both the storage and processing time can be highly optimized with that in mind. The results of the preprocessing are sparse matrixes, and hence the optimization is feasible. The Chamfer distance, for instance, can compute just the distances between pixels that are not background (background pixels are the majority).

Although the highest true positive rate obtained was 17.02% with the entire database, the approach can reach up to maximum TP (all recognitions are correct) if the dataset is relatively small (up to around 20 users that have 12 images each, resulting on 360 comparisons). Furthermore, the recognition rate can also be substantially improved if more than one sample is used to recognize the person. That is, asking the user to be scanned at least twice, instead of once. That would certainly impact negatively on the processing, but the recognition rates would improve.

A future work would be to test the performance of grey-image oriented similarity measures. Even though it cannot be faster than this approach, it could potentially obtain better recognition rates if the images are registered and compared (which is also a very slow process). Another possible direction would be to face the problem using machine learning algorithms.

Table IV: Comparison of obtained results.

| Works | True Positive (%) | Accuracy (%) |
|---|---|---|
| This work | 17.023 | 99.99 |
| Zhang et al. | 15.15 | - |
| Le-qing | - | 96.91 |
| Guru et al. | - | 92.24 |

## Acknowledgment

The authors are grateful to CNPq, CAPES and FAPERJ for the financial suppport of this work.